\UseRawInputEncoding
\documentclass[manuscript,nonacm]{acmart}
\setcopyright{none}
\renewcommand\footnotetextcopyrightpermission[1]{}

\title{NEURON: A Neuro-symbolic System for Grounded Clinical Explainability}
\author{Anuradha Chandrasekaran}
\affiliation{
  \institution{University of North Texas}
  \city{Denton}
  \state{TX}
  \country{USA}
}
\affiliation{
  \institution{Texas Tech University Health Sciences Center}
  \city{Lubbock}
  \state{TX}
  \country{USA}
}
\email{anuradhachandrasekaran@my.unt.edu}

\author{Dimitrios Zikos}
\affiliation{
  \institution{Texas Tech University Health Sciences Center}
  \city{Lubbock}
  \state{TX}
  \country{USA}
}
\email{dzikos@ttuhsc.edu}

\author{Mutlu Mete}
\affiliation{
  \institution{University of North Texas}
  \city{Denton}
  \state{TX}
  \country{USA}
}
\email{mutlu.mete@unt.edu}

\author{Alan Pang}
\affiliation{
  \institution{Texas Tech University Health Sciences Center}
  \city{Lubbock}
  \state{TX}
  \country{USA}
}
\email{alan.pang@ttuhsc.edu}

\author{Brady D. Lund}
\affiliation{
  \institution{University of North Texas}
  \city{Denton}
  \state{TX}
  \country{USA}
}
\email{brady.lund@unt.edu}

\author{Kewei Sha}
\affiliation{
  \institution{University of North Texas}
  \city{Denton}
  \state{TX}
  \country{USA}
}
\email{kewei.sha@unt.edu}

\usepackage{tabularx}
\usepackage{booktabs}
\usepackage{graphicx}
\usepackage{placeins}
\usepackage{capt-of} 
\usepackage{tcolorbox}
\tcbuselibrary{breakable,listings,skins}
\usepackage{listings}
\usepackage{xcolor}
\usepackage{tikz}
\usepackage{xcolor}
\usepackage{pifont}
\usepackage[utf8]{inputenc}
\usepackage[T1]{fontenc}
\usepackage{placeins}
\usepackage[skip=3pt]{caption}
\usepackage{float}
\usepackage{dblfloatfix}
\usepackage{cuted}
\newcommand{\cmark}{\textcolor{green!60!black}{\ding{51}}}
\newcommand{\cmarkg}{\textcolor{black!45}{\ding{51}}} 
\newcommand{\xmark}{\textcolor{red!70!black}{\ding{55}}}

\lstdefinestyle{promptstyle}{
  basicstyle=\ttfamily\scriptsize,
  breaklines=true,
  columns=fullflexible,
  frame=none,
  showstringspaces=false,
  keepspaces=true,
  xleftmargin=0pt,
  xrightmargin=0pt,
  aboveskip=0pt,
  belowskip=0pt,
  resetmargins=true,
  breakindent=0pt      
}
\begin{document}

\begin{abstract}
Clinical AI adoption is hindered by the black-box/grey-box nature of high-performing models, which lack the ontological grounding and narrative transparency required for professional-level explainability. We present NEURON, a neuro-symbolic system designed to enhance both predictive reliability and clinical interpretability. NEURON integrates SNOMED CT ontology-informed structural representations with machine learning models to bridge the gap between raw data and medical nomenclature. To facilitate human-aligned interaction, the system utilizes a Retrieval-Augmented Generation (RAG) grounded Large Language Model (LLM) layer to synthesize SHAP feature attributions and patient-specific clinical notes into coherent, natural-language explanations. Validated on the MIMIC-IV dataset for Acute Heart Failure mortality prediction, NEURON improved the AUC from 0.71–0.74 to 0.74–0.77 and substantially outperformed SHAP-based explanations in human-aligned metrics (0.807 vs. 0.558). Our results demonstrate that NEURON offers a robust, scalable engineering solution for deploying trustworthy, human-centered connected health applications.
\end{abstract}

\maketitle
\begin{center}
\textit{Accepted at IEEE CHASE 2026. This is the author’s preprint version.}
\end{center}

\section{Introduction}
The use of AI is growing in medicine across multiple areas, including diagnosis, treatment, dose optimization, and drug monitoring \cite{alowaisRevolutionizingHealthcareRole2023a, brigantiArtificialIntelligenceMedicine2020}. The algorithms are widely used for their high accuracy. Predictive performance is crucial in healthcare because accurate risk estimates can enable prioritization of care and timely interventions \cite{britsch_interpretable_2025, thorsen2020dynamic, weissman_inclusion_2018}. However, despite strong performance, the machine learning (ML) models currently used do not support meaningful clinical sensemaking for users with diverse roles and expertise, as the reasoning behind their accurate results remains opaque \cite{adadiPeekingBlackboxSurvey2018b,aliExplainableArtificialIntelligence2023b}. Explainable AI (XAI) refers to techniques used to unbox AI algorithms and understand the reasoning behind their decisions \cite{adadiPeekingBlackboxSurvey2018b}. 

Several reasons prompt this call to integrate explainability within AI-based models in the medical field. Under the General Data Protection Regulation (GDPR), individuals must be informed about the underlying decision-making and the logic used, its importance, and any consequences for the user \cite{schneeberger_european_2020}. It is essential to understand how a model arrives at a decision, especially in healthcare, where incorrect decisions that are acted upon can have life-or-death consequences \cite{adadiPeekingBlackboxSurvey2018b,longoExplainableArtificialIntelligence2024,saeedExplainableAIXAI2021}.

Currently, several XAI algorithms use various approaches to explain ML models, particularly those that are not easily interpretable. These approaches can be categorized as model-specific or model-agnostic: model-specific techniques depend on the model under consideration, whereas model-agnostic techniques do not require access to the model's internal workings. Another categorization is between local and global interpretability. Local interpretability refers to explaining the predictions for a specific data point, while global refers to explaining the overall model behavior \cite{adadiPeekingBlackboxSurvey2018b,arrietaExplainableArtificialIntelligence2019a,dwivediExplainableAIXAI2022}. One well-known and widely used XAI technique is SHAP (SHapley Additive exPlanations) \cite{lundbergUnifiedApproachInterpreting2017,ortigossaEXplainableArtificialIntelligence2024}, a model-agnostic feature-attribution method that uses graphs and other visual modalities to show which features contribute most to the final output. It can be used for both local and global interpretability \cite{dwivediExplainableAIXAI2022,ortigossaEXplainableArtificialIntelligence2024}. Explainability techniques like SHAP still provide only static feature-attribution visualizations that are difficult for clinicians to interpret \cite{lee_prompt_2025,longoExplainableArtificialIntelligence2024, saeedExplainableAIXAI2021}. Furthermore, these techniques do not incorporate clinical knowledge, nor do they offer a mechanism for interactive or narrative reasoning. This hinders the ability to trace predictions back to clinically grounded knowledge, thereby limiting trust and adoption of ML in healthcare settings \cite{confalonieriMultipleRolesOntologies2024,lee_prompt_2025,longoExplainableArtificialIntelligence2024, saeedExplainableAIXAI2021}.

This study proposes developing a framework, named NEURON (Neuro-symbolic Understanding with Retrieval Oriented Narration), which not only adds ontology-informed representations to a sub-symbolic ML model to increase performance and thus converting the design to neuro-symbolic but also includes a RAG-grounded LLM layer to generate natural-language explanations of established clinical knowledge, model-derived drivers, and patients' stay-specific clinical notes. Our main contribution, NEURON, is a unified neuro-symbolic framework that jointly improves both prediction performance and explainability requirements in clinical AI systems.

The experiments are built on in-hospital mortality prediction in acute heart failure (AHF) patients using the MIMIC-IV dataset, by augmenting standard tabular clinical features with SNOMED CT (Systematized Nomenclature of Medicine Clinical Terms) ontology-derived representations. The ontology integration is handled non-trivially, since we do not treat SNOMED codes as one-hot features. We create three distinct representations: TF-IDF-weighted Node2Vec embeddings, explicit counts of ontology categories, and SHAP feature attributions. The SHAP model drivers, structural ontology representations, clinical notes grounded in patient context, and manually curated clinical knowledge are all provided as input to the RAG layer. This enables explanations that account for multiple sources of information and provide a comprehensive picture of a patient's condition, thereby enhancing understandability for healthcare providers. Furthermore, we use human-aligned and computational metrics to compare SHAP outputs and the narrative explanations generated by the LLM. 

This paper is organized as follows. Section II presents related work on explainable AI and neuro-symbolic approaches in healthcare. Section III describes the design and rationale of the proposed NEURON framework. Section IV outlines the experiments (data sources, cohort construction, feature representations, data transformation, ontology integration, and predictive modeling). Section V reports performance results across sub-symbolic and neuro-symbolic configurations. Section VI details the explanation-generation pipeline that uses SHAP results and a RAG-grounded LLM, along with the evaluation metrics. Section VII compares explanation quality using computational and human-aligned metrics. Section VIII discusses limitations and challenges. Section IX concludes and Section X outlines future work. 

\section{Related Work}
Symbolic AI consists of a rule-based system where each piece of knowledge is embedded into the model itself, and could be traversed using logical, predefined rules \cite{garcezNeurosymbolicAI3rd2020a, hossainStudyNeuroSymbolicArtificial2025a, shethNeuroSymbolicKnowledgeGroundedPlanning2025b}. This makes the results highly interpretable because we can trace their underlying reasoning. However, it is strictly limited to manual programming and struggles to learn from the data itself to produce new rules. Sub-symbolic AI, by contrast, can learn complex, high-dimensional rule mappings directly from the data, making it highly scalable and adaptive \cite{hossainStudyNeuroSymbolicArtificial2025a, piplaiKnowledgeEnhancedNeurosymbolicArtificial2023}. However, its decision-making reasoning process is opaque. For clinical decision-making, this may mean a model with high predictive performance (AUC > 0.8) lacking a clinician-understandable justification for its output.

The XAI literature has proposed several methods to address this issue, including popular XAI techniques like SHAP that assign feature importance to model prediction behavior \cite{ arrietaExplainableArtificialIntelligence2019a,lundbergUnifiedApproachInterpreting2017,ortigossaEXplainableArtificialIntelligence2024, saeedExplainableAIXAI2021}. Reviews specific to healthcare point to the black-box nature of the internal decision logic of AI models used in healthcare decision-making \cite{amannExplainabilityArtificialIntelligence2020, holzingerCausabilityExplainabilityArtificial2019, rubaiyatReviewExplainableArtificial2023a}. They further identify two shortcomings with techniques currently being used: (a) explanations are static and lack the ability to facilitate causality, counterfactuals, and user interpretability \cite{holzingerCausabilityExplainabilityArtificial2019, longoExplainableArtificialIntelligence2024, naveedOverviewEmpiricalEvaluation2024, saeedExplainableAIXAI2021}; (b) they are not designed or evaluated with human cognition in mind \cite{alhasanEnhancingAIExplainability2025, longoExplainableArtificialIntelligence2024}. This may consequently not always yield explanations that are helpful to end users who lack the expertise to understand them \cite{lee_prompt_2025,hossainStudyNeuroSymbolicArtificial2025a,piplaiKnowledgeEnhancedNeurosymbolicArtificial2023}.

The third wave of AI proposes combining symbolic knowledge structures with sub-symbolic AI to produce systems that are both adaptive and interpretable, which enables trustworthy decision-making \cite{gaurBuildingTrustworthyNeuroSymbolic2023a, shethNeurosymbolicArtificialIntelligence2023b}. Medical ontologies formalize clinical concepts and their hierarchical relationships, making them well-suited for this purpose \cite{confalonieriMultipleRolesOntologies2024}. Neuro-symbolic methods that incorporate clinical ontologies can create more causal and faithful explanations that are better aligned with established clinical knowledge \cite{hossainStudyNeuroSymbolicArtificial2025a,longoExplainableArtificialIntelligence2024, shethNeuroSymbolicKnowledgeGroundedPlanning2025b}. 

LLMs add the capability of translating machine learning explanations into audience-specific natural language narratives that can be reasoned with \cite{lee_prompt_2025, zytekLLMsXAIFuture2024}. Further research shows that participants prefer narrative explanations to plot-based ones \cite{zytekLLMsXAIFuture2024}. However, in the medical context, the risk of hallucination, where the system generates a plausible but incorrect statement, is a safety concern. Hence, using a RAG layer to ground LLMs in domain knowledge helps maintain factual accuracy \cite{lewis2020retrieval,nehaRetrievalAugmentedGenerationRAG2025}.

Accurate models that predict the risk of in-hospital mortality in patients with acute heart failure are essential for risk stratification, resource allocation, early intervention, and patient management \cite{chenMachineLearningbasedInhospital2023, huang_prediction_2024,  liMachineLearningInhospital2025, misumi_derivation_2023}. This motivates us to develop models that reliably predict outcomes, support timely clinical decision-making, and offer clinically grounded interpretability.

\subsection{Research Gap}
Research in this field has addressed topics in isolation. Ontology-grounded RAG has been shown to reduce hallucinations and improve response accuracy across several LLMs. This has been demonstrated in recent ontology-aware RAG systems and in healthcare-focused surveys \cite{nehaRetrievalAugmentedGenerationRAG2025, sharmaOGRAGOntologyGroundedRetrievalAugmented2024}. Clinician-facing studies that convert SHAP attribution scores into natural-language narratives received positive ratings for categories such as helpfulness and usefulness \cite{lee_prompt_2025}, but they fail to ground LLMs in existing structured clinical knowledge (e.g., ontologies/KBs/guidelines). Neuro-symbolic frameworks have been proposed that combine knowledge graphs with ML algorithms \cite{shethNeurosymbolicArtificialIntelligence2023b}, and further research has used hand-constructed knowledge structures with clinically defined thresholds to feed into an ML pipeline without losing predictive performance \cite{sirocchi_medical-informed_2024}. However, to the best of our knowledge, no prior study presents a controlled comparison of sub-symbolic and neuro-symbolic pipelines on the same clinical cohort, where (a) SHAP attribution is directly coupled with the RAG-grounded LLM, which further combines multiple sources of information, such as patients’ stay-specific ontological representations, unstructured clinical notes, and a manually curated knowledge base, (b) explanation quality is evaluated for both SHAP and narrative outputs across computational and human-aligned metrics. Table~\ref{tab:neuron_comparison_newschema} summarizes the unique contributions of NEURON compared with existing work using MIMIC and similar contextual data.

\begin{table*}[htbp]
\centering
\footnotesize
\caption{Comparison of prediction, ontology integration, XAI attribution, LLM narrative explanation, and interpretability evaluation grouped by methodological paradigm.}
\label{tab:neuron_comparison_newschema}
\renewcommand{\arraystretch}{1.25}
\setlength{\tabcolsep}{3pt}
\begin{tabularx}{\textwidth}{p{2.0cm} X c c c c c c c c}
\hline
\textbf{Study} &
\textbf{Dataset} &
\textbf{Pred ML?} &
\textbf{Ont$\rightarrow$Pred?} &
\textbf{XAI?} &
\textbf{LLM Narr?} &
\textbf{Ont$\rightarrow$LLM?} &
\textbf{SHAP$\rightarrow$LLM?} &
\textbf{Metrics?} &
\textbf{Multimodal?} \\
\hline

\textbf{NEURON (proposed)} &
MIMIC-IV (AHF ICU) &
\cmark &
\cmark &
\cmark &
\cmark &
\cmark &
\cmark &
\cmark &
\cmark \\

\hline
\multicolumn{10}{l}{\textit{A. Standard ML/DL prediction + post-hoc SHAP/XAI:}} \\
\hline

Chen et al. \cite{chenMachineLearningbasedInhospital2023} &
MIMIC-IV + eICU &
\cmark &
\xmark &
\cmark &
\xmark &
\xmark &
\xmark &
\xmark &
\xmark \\

Huang et al. \cite{huang_prediction_2024} &
MIMIC-IV (AHF ICU) &
\cmark &
\xmark &
\cmark &
\xmark &
\xmark &
\xmark &
\xmark &
\xmark \\

Xie et al. \cite{xie_development_2024} &
MIMIC-IV &
\cmark &
\xmark &
\cmark &
\xmark &
\xmark &
\xmark &
\xmark &
\xmark \\

Li et al. \cite{liMachineLearningInhospital2025} &
MIMIC-IV (AHF ICU) &
\cmark & \xmark & \xmark & \xmark & \xmark & \xmark & \xmark & \xmark \\

Mesinovic et al. \cite{mesinovic_explainable_2025} &
MIMIC-IV + eICU &
\cmark &
\xmark &
\cmark &
\xmark &
\xmark &
\xmark &
\cmark  &
\xmark \\

Zheng et al. \cite{zheng_explainable_2025} &
MIMIC-IV (MV ICU) &
\cmark &
\xmark &
\cmark &
\xmark &
\xmark &
\xmark &
\xmark &
\xmark \\

\hline
\multicolumn{10}{l}{\textit{B. Ontology/KG injected into predictor:}} \\
\hline

Niu et al. \cite{niuFusionSequentialVisits2022} &
MIMIC-III &
\cmark &
\cmark &
\xmark &
\xmark &
\xmark &
\xmark &
\xmark &
\xmark \\

Wang \& Li \cite{wang_integrating_2025} &
MIMIC-IV &
\cmark &
\cmark &
\xmark &
\xmark &
\xmark &
\xmark &
\xmark &
\cmarkg \\

Jiang et al. \cite{jiang_reasoning-enhanced_2025} &
MIMIC-III/IV &
\cmark &
\cmark &
\xmark &
\cmark &
\cmark &
\xmark &
\cmark &
\cmark \\

\hline
\multicolumn{10}{l}{\textit{C. LLM-based narrative explanation }} \\
\hline

Lee et al. \cite{lee_prompt_2025} &
MIMIC-III &
\cmark &
\xmark &
\cmark &
\cmark &
\xmark &
\cmark &
\cmark &
\xmark \\

\hline
\multicolumn{10}{l}{\textit{D. Contextual non-MIMIC frameworks:}} \\
\hline

Sharma et al. \cite{sharmaOGRAGOntologyGroundedRetrievalAugmented2024} &
Context &
\xmark &
\xmark &
\xmark &
\cmark &
\cmark &
\xmark &
\cmark &
\cmarkg \\

Feng et al. \cite{feng_ontologyrag_2025} &
Context &
\xmark &
\xmark &
\xmark &
\cmark &
\cmark &
\xmark &
\cmarkg &
\cmarkg \\

Hur et al. \cite{hur_comparison_2025} &
Context &
\cmark &
\xmark &
\cmark &
\cmarkg &
\xmark &
\cmarkg &
\cmark &
\xmark \\

\hline
\end{tabularx}

\vspace{0.3em}
\begin{minipage}{0.98\textwidth}
\footnotesize
\raggedright
\textbf{Legend:} \cmark=yes; \xmark=no; \cmarkg=partial/indirect. \
\textbf{Dataset abbreviations:} 
AHF=acute heart failure; MV=mechanically ventilated; 
eICU=eICU Collaborative Research Database. \
\textbf{Pred ML?}: clinical outcome prediction with ML/DL/LLM. \;
\textbf{Ont$\rightarrow$Pred?}: ontology/KG injected into predictor. \;
\textbf{XAI?}: SHAP/LIME attribution. \;
\textbf{LLM Narr?}: narrative explanation generated. \;
\textbf{Ont$\rightarrow$LLM?}: ontology grounding for LLM. \;
\textbf{SHAP$\rightarrow$LLM?}: SHAP drivers passed to LLM. \;
\textbf{Metrics?}: explanation quality evaluated. \;
\textbf{Multimodal?}: uses notes/KB/text beyond numeric tabular data.
\end{minipage}
\end{table*}

\section{Design of NEURON}
The design of the NEURON framework is motivated by our intention to improve predictive performance and explanation quality across heterogeneous black-box ML models that are hard to interpret. High-performing ML models often fail to provide explanations that clinicians can interpret and act upon \cite{senguptaHowCliniciansThink2026a, tonekaboniWhatCliniciansWant2019}. NEURON improves predictive performance and generates narrative explanations derived from multiple sources and modes of clinical information, including structured tabular variables, ontology-informed representations, and unstructured patient context. 

Our design targets ML model families, such as MLP, XGBoost, Autoencoder, and RBF SVM, which are widely used in healthcare prediction tasks due to their high predictive performance but whose internal reasoning is not easily interpretable \cite{chenMachineLearningbasedInhospital2023, guido_overview_2024, HAKKOUM2024107829, liMachineLearningInhospital2025, miotto_deep_2016}. This allows us to see whether the proposed neuro-symbolic framework consistently improves the predictive performance of these algorithms trained on the same MIMIC-IV cohort. NEURON integrates an ontology layer that provides structural representations of established clinical knowledge to both ML models for prediction and the narrative explanation layer, constraining the LLM's retrieval of relevant concepts and their relationships before producing text, thereby reducing hallucinations and improving factual consistency. 

NEURON uses SNOMED CT to construct an “is-a” knowledge graph to represent structured clinical relationships. SNOMED CT is a clinical terminology used by healthcare researchers worldwide to facilitate the storage and analysis of clinical data \cite{NLM_SNOMED_Sep2025}. 
This SNOMED ontology layer acts as a blueprint for organizing facts and their relationships within the healthcare domain. This provides the structured ontology context that neuro-symbolic and RAG-based pipelines use to create knowledge graphs, is-a hierarchies, and admission-level mappings to SNOMED codes \cite{pengKnowledgeGraphsOpportunities2023,purohitEnhancingMedicalKnowledge2025,sharmaOGRAGOntologyGroundedRetrievalAugmented2024}. Node2Vec \cite{grover2016node2vec} embeddings are created based on this graph, and admission-level embeddings are further generated using TF-IDF weighted pooling, which results in one fixed 32-dimensional vector per admission. Level 1 of the ontology is a curated set of SNOMED CT concepts, while level 2 is automatically selected, and a bag of ontology features mapped to these two levels is derived for each admission. The tabular features, SNOMED-derived embeddings, and ontology counts collectively constitute the input to the neuro-symbolic framework, which is run for each ML model. 

\begin{figure}[t]
\centering
\includegraphics[width=1.00\linewidth,height=5cm,keepaspectratio]{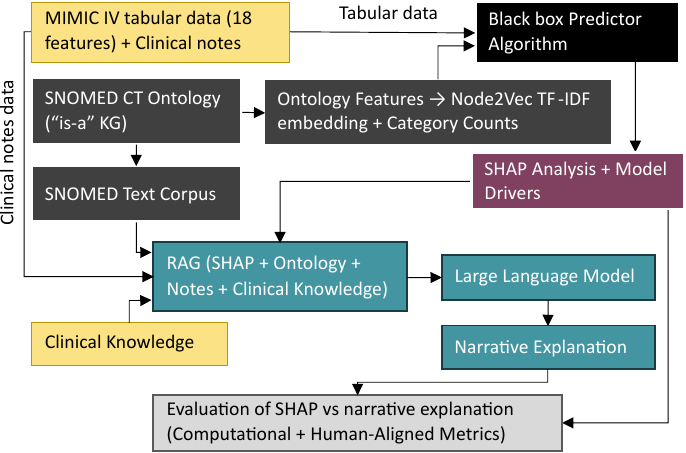}
\caption{Overall design of the NEURON framework.}
  \label{fig:neuron_overview}
\end{figure}

SHAP is used to provide a unified feature-attribution approach across each model. SHAP explanations provide feature attribution scores that are typically static, not grounded in clinical knowledge, and difficult to interpret \cite{confalonieriMultipleRolesOntologies2024,kaurSensibleAIReimagining2022a,lee_prompt_2025, longoExplainableArtificialIntelligence2024, saeedExplainableAIXAI2021}. Hence, the need to create a neuro-symbolic framework and compare it with ML algorithms that are widely used in this field for both predictive performance as well as quality of explanations, which are explicitly grounded in clinical knowledge \cite{shethNeuroSymbolicKnowledgeGroundedPlanning2025b}. The patient’s SHAP profile, clinical ontology structure, clinical notes, and clinical knowledge detailing thresholds, directionality, and interpretation are used by the RAG layer to generate a narrative explanation that supports clinical sensemaking by providing a holistic picture of the patient’s condition. Figure~\ref{fig:neuron_overview} showcases the overall design of the NEURON framework.

\section{Experiments}
\subsection{Data Source and Cohort Construction}
We chose AHF as the clinical scenario since it has a high incidence rate in the US, and around 10\% to 51\% of the patients suffering from HF are admitted to the intensive care unit (ICU), with an in-hospital mortality rate of 10.6\% \cite{liMachineLearningInhospital2025}. Additionally, this medical condition has been well studied for predicting mortality risk, and the necessary data are consistently available as structured predictors (e.g., vital signs and routine laboratory tests) \cite{blaziak_artificial_2022,nadarajah_prediction_2023}. This makes AHF particularly suitable for reproducible modeling and explanation studies using EHR/ICU data. The MIMIC-IV database (V3.1) is a de-identified publicly available healthcare data repository managed by the Massachusetts Institute of Technology (MIT) and approved by the IRB boards of both Beth Israel Deaconess Medical Center and MIT \cite{goldberger2000physiobank, johnsonMIMICIVFreelyAccessible2023,PhysioNet-mimiciv-3.1}. This data repository consists of approximately 94,400 ICU stays across 65,366 unique individuals, including details such as vital signs, laboratory results, oxygen saturation, medications, and ICD codes for diagnoses and procedures. Free-text clinical documentation, such as notes, is available via linked resources (MIMIC-IV-Note). PostgreSQL and Structured Query Language (SQL) were used to extract and store data in a relational database. Views specific to the AHF cohort consisted of patients over 18 years of age and admitted to the ICU with either ICD-9 (42821, 42831, 42841, 42823, 42833, and 42843) or ICD-10 codes (I5021, I5031, I5041, I50811, I5023, I5033, I5043, and I50813). ICU admission served as the additional clinical acuity criterion. 

\subsection{Feature Representation}
We selected 18 features for AHF mortality prediction and created a separate PostgreSQL view to store them. We narrowed the set of 18 features based on the paper by Li et al. \cite{liMachineLearningInhospital2025}, who initially identified 143 variables and then applied LASSO regression to select these 18 significant variables. The variables we selected are age and other clinical features measured within 24 hours of ICU admission (Table~\ref{tab:descriptives}). 

\begin{table*}[htbp]
\centering
\small
\caption{Baseline characteristics of the study cohort before data preprocessing.}
\label{tab:descriptives}
\footnotesize
\setlength{\tabcolsep}{3pt}  
\renewcommand{\arraystretch}{1.05} 

\begin{tabularx}{\textwidth}{>{\raggedright\arraybackslash}X c c c c c c c}
\toprule
 & \textbf{Total} & \textbf{Alive} & \textbf{Died} & \textbf{p-value} & \textbf{Training} & \textbf{Validation} & \textbf{p-value} \\
 & n=13{,}659 & n=11{,}556 & n=2{,}103 & Alive vs Died & n=9{,}597 & n=4{,}062 & Train vs Val \\
\midrule
Age, mean (SD) & 69.72 (13.80) & 69.12 (13.96) & 72.99 (12.42) & $<0.001$ & 69.62 (13.86) & 69.94 (13.67) & 0.228 \\
WBC$_{\text{min}}$, mean (SD) & 11.10 (6.95) & 10.82 (6.63) & 12.63 (8.37) & $<0.001$ & 11.06 (7.08) & 11.18 (6.63) & 0.084 \\
Anion gap$_{\text{min}}$, mean (SD) & 13.62 (3.80) & 13.34 (3.50) & 15.20 (4.88) & $<0.001$ & 13.64 (3.82) & 13.58 (3.76) & 0.608 \\
Anion gap$_{\text{max}}$, mean (SD) & 16.09 (4.56) & 15.68 (4.14) & 18.41 (5.90) & $<0.001$ & 16.13 (4.60) & 15.99 (4.47) & 0.202 \\
BUN$_{\text{min}}$, mean (SD) & 36.47 (24.98) & 34.61 (23.68) & 46.94 (29.18) & $<0.001$ & 36.84 (25.56) & 35.60 (23.55) & 0.224 \\
INR$_{\text{min}}$, mean (SD) & 1.65 (0.92) & 1.61 (0.88) & 1.87 (1.08) & $<0.001$ & 1.64 (0.90) & 1.67 (0.96) & 0.090 \\
INR$_{\text{max}}$, mean (SD) & 1.85 (1.28) & 1.78 (1.19) & 2.21 (1.62) & $<0.001$ & 1.84 (1.26) & 1.87 (1.32) & 0.381 \\
PTT$_{\text{min}}$, mean (SD) & 37.46 (18.89) & 36.82 (18.20) & 40.89 (21.89) & $<0.001$ & 37.43 (19.15) & 37.55 (18.26) & 0.128 \\
Urine Output, mean (SD) & 1830.54 (1369.94) & 1923.05 (1379.19) & 1294.93 (1180.60) & $<0.001$ & 1833.04 (1365.91) & 1824.64 (1379.57) & 0.737 \\
Heart rate, mean (SD) & 85.93 (16.76) & 85.34 (16.47) & 89.19 (17.89) & $<0.001$ & 85.99 (16.72) & 85.80 (16.85) & 0.602 \\
SBP, mean (SD) & 114.31 (16.71) & 115.40 (16.57) & 108.34 (16.19) & $<0.001$ & 114.37 (16.76) & 114.18 (16.57) & 0.686 \\
DBP$_{\text{min}}$, mean (SD) & 44.71 (11.61) & 45.37 (11.38) & 41.09 (12.21) & $<0.001$ & 44.82 (11.64) & 44.47 (11.53) & 0.230 \\
Respiratory rate, mean (SD) & 20.43 (3.85) & 20.26 (3.73) & 21.39 (4.36) & $<0.001$ & 20.39 (3.84) & 20.52 (3.87) & 0.056 \\
SpO$_2{}_{\text{min}}$, mean (SD) & 90.10 (6.81) & 90.53 (5.88) & 87.71 (10.27) & $<0.001$ & 90.09 (6.87) & 90.11 (6.68) & 0.646 \\
Dobutamine, n (\%) & 579 (4.8\%) & 406 (3.9\%) & 173 (9.9\%) & $<0.001$ & 397 (4.7\%) & 182 (5.0\%) & 0.469 \\
Dopamine, n (\%) & 586 (4.9\%) & 417 (4.0\%) & 169 (9.6\%) & $<0.001$ & 406 (4.8\%) & 180 (5.0\%) & 0.733 \\
Norepinephrine, n (\%) & 2725 (22.6\%) & 2007 (19.5\%) & 718 (40.9\%) & $<0.001$ & 1918 (22.7\%) & 807 (22.3\%) & 0.634 \\
Phenylephrine, n (\%) & 1349 (11.2\%) & 1119 (10.8\%) & 230 (13.1\%) & 0.006 & 969 (11.5\%) & 380 (10.5\%) & 0.127 \\
\bottomrule
\end{tabularx}
\scriptsize \textbf{Abbreviations:} SD, standard deviation; WBC, white blood cell count; BUN, blood urea nitrogen; INR, international normalized ratio; PTT, partial thromboplastin time; SBP, systolic blood pressure; DBP, diastolic blood pressure; SpO$_2$, peripheral oxygen saturation. 
\end{table*}

\subsection{Data Transformation}
Observations with missing outcome labels were removed. The train-validation split was performed at the subject level, ensuring that all stays from a given patient appeared in only one split. For each feature configuration (tabular-only, tabular+KG embeddings, and full neuro-symbolic), observations with more than 30\% missing predictor values were excluded. The remaining data was intersected at the stay identifier level (training n = 9,294; validation n = 3,951) to ensure that all ML models were trained and evaluated on the same set of ICU stays. All models used a consistent imputation strategy for the remaining observations, in which numeric fields were imputed with the median. Feature scaling was applied only to models that are scale-sensitive: the MLP and RBF-SVM use z-score standardization (StandardScaler) after median imputation. The autoencoder model applies min–max normalization internally prior to training and inference. XGBoost, which is a tree-based model, was trained on unscaled features since their threshold-based split decisions are dependent on feature ordering and are invariant monotonic transformations of scale \cite{molnar2025}.

\subsection{Knowledge Graph and Embeddings}
The 18 features form the tabular baseline feature vector for our sub-symbolic framework \cite{liMachineLearningInhospital2025}. We first constructed a full SNOMED CT graph using concepts, descriptions, and relationships (is-a). We then automatically discovered level 2 ontology anchor categories from level 1 via descendant traversal and organized them into multi-level ontology categories.  Vector embeddings were computed for this SNOMED graph using a custom Node2Vec \cite{grover2016node2vec} implementation, providing a vector representation for each SNOMED concept. Because each hospital admission was associated with multiple SNOMED codes, we generated one fixed-length vector per admission using TF-IDF-weighted pooling of SNOMED concept embeddings to represent them consistently across all hospital admissions and aligned these embeddings with ICU stays for modeling. In addition, we computed aggregated counts of ontology features at levels 1 and 2. When embedding or ontology-derived features were absent for a given stay, they were filled with zeros to ensure that we represented the lack of ontology features rather than data missingness.

\subsection{Predictive Models}
\begin{figure*}[t]
  \centering
  \includegraphics[width=\textwidth]{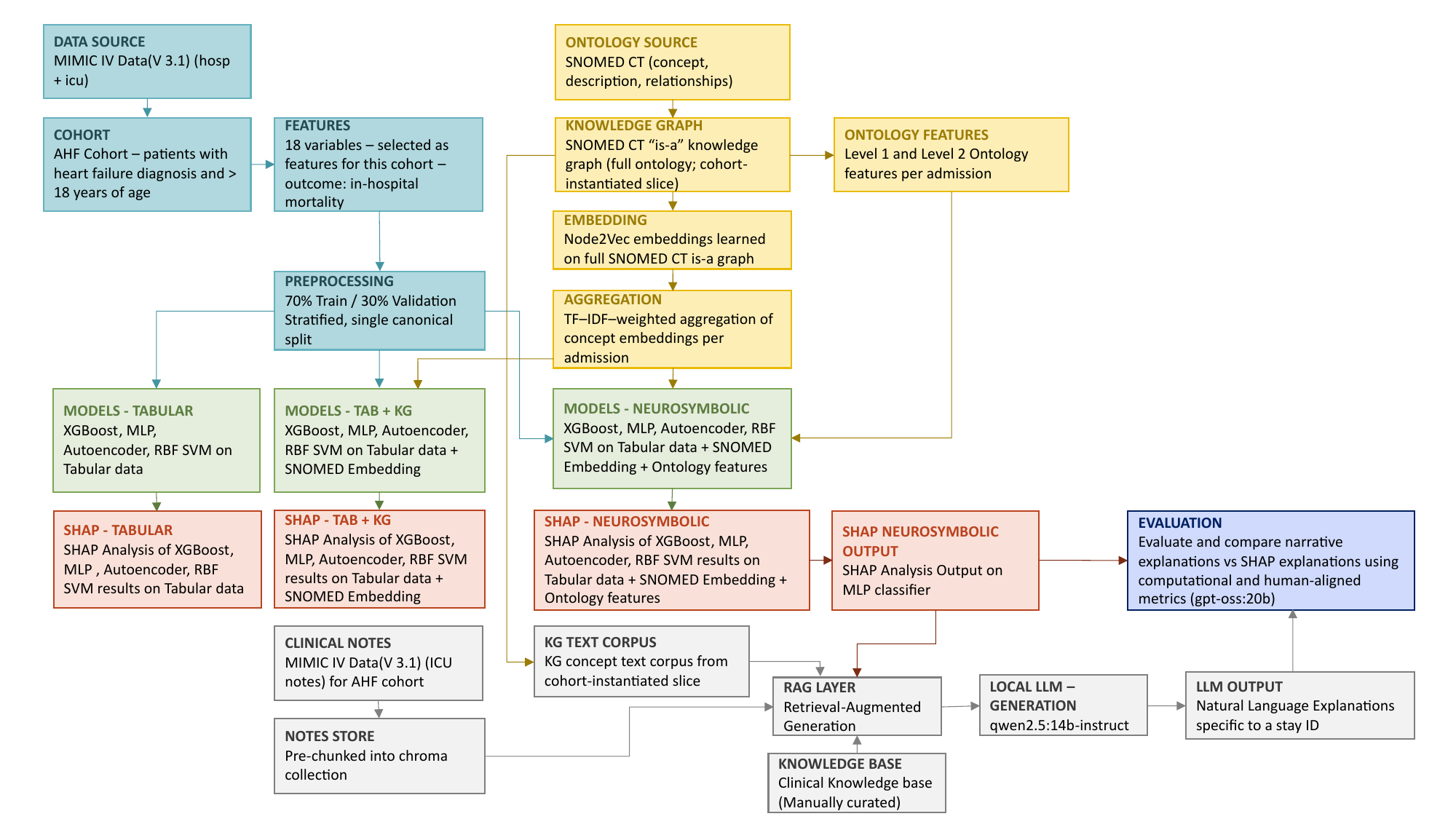}
  \caption{Overview of the neuro-symbolic modeling and explanation pipeline for in-hospital mortality prediction in AHF.}
  \label{fig:pipeline}
\end{figure*}

Python was used to implement the ML algorithms (XGBoost, MLP classifier, RBF SVM, and Autoencoder) to predict in-hospital mortality for the AHF cohort. Prediction was formulated at the ICU stay level under three progressively enriched feature representations: (i) tabular baseline features, (ii) tabular features plus SNOMED graph embeddings, and (iii) a full neuro-symbolic configuration combining embeddings with ontology-derived category counts. Figure~\ref{fig:pipeline} presents the entire flow of data and processes, providing an overview of the neuro-symbolic modeling and explanation pipeline. This enabled us to compare predictive performance across these algorithms under three different configurations. Table~\ref{tab:performance} presents our performance results for the algorithms and their configurations.

\begin{table*}[htbp]
\centering
\caption{Predictive performance of ML models for in-hospital mortality prediction under three feature configurations.}
\label{tab:performance}
\small
\setlength{\tabcolsep}{2pt}
\renewcommand{\arraystretch}{1.10}
\begin{tabular}{lccccccc}
\toprule
\textbf{Model} & \textbf{AUC} & \textbf{95\% CI} & \textbf{Accuracy} & \textbf{Sensitivity} & \textbf{Precision} & \textbf{F1} & \textbf{Specificity} \\
\midrule
XGB (Tabular)                   & 0.73 & 0.71--0.75 & 0.85 & 0.20 & 0.57 & 0.30 & 0.97 \\
XGB (Tabular + KG emb)          & 0.75 & 0.74--0.77 & 0.85 & 0.17 & 0.57 & 0.26 & 0.98 \\
XGB (NeuroSymbolic)             & 0.77 & 0.76--0.79 & 0.85 & 0.18 & 0.56 & 0.27 & 0.97 \\
\midrule
MLP (Tabular)                   & 0.74 & 0.72--0.76 & 0.73 & 0.60 & 0.31 & 0.41 & 0.76 \\
MLP (Tabular + KG emb)          & 0.75 & 0.73--0.77 & 0.68 & 0.68 & 0.28 & 0.39 & 0.68 \\
MLP (NeuroSymbolic)             & 0.76 & 0.74--0.78 & 0.73 & 0.61 & 0.30 & 0.41 & 0.75 \\
\midrule
Autoencoder (Tabular)           & 0.71 & 0.69--0.73 & 0.70 & 0.60 & 0.28 & 0.38 & 0.72 \\
Autoencoder (Tabular + KG emb)  & 0.75 & 0.73--0.78 & 0.74 & 0.59 & 0.31 & 0.41 & 0.77 \\
Autoencoder (NeuroSymbolic)     & 0.74 & 0.71--0.76 & 0.72 & 0.60 & 0.30 & 0.40 & 0.75 \\
\midrule
RBF-SVM (Tabular)               & 0.73 & 0.70--0.75 & 0.84 & 0.07 & 0.44 & 0.12 & 0.98 \\
RBF-SVM (Tabular + KG emb)      & 0.75 & 0.74--0.77 & 0.85 & 0.13 & 0.51 & 0.21 & 0.98 \\
RBF-SVM (NeuroSymbolic)         & 0.76 & 0.74--0.78 & 0.85 & 0.18 & 0.54 & 0.27 & 0.97 \\
\bottomrule
\end{tabular}
\vspace{0.5em}
\scriptsize

\textbf{Abbreviations:} 
AUC, area under the ROC curve; 
CI, confidence interval; 
KG, knowledge graph; 
XGB, extreme gradient boosting; 
MLP, multilayer perceptron; 
SVM, support vector machine; 
RBF, radial basis function; 
RBF-SVM, radial basis function support vector machine; 
AE, autoencoder.
\end{table*}

\section{Predictive Performance Results}
Performance was evaluated at the ICU stay level, and confidence intervals were obtained via bootstrap resampling of validation stays. As shown in Table~\ref{tab:performance}, performance generally improved as SNOMED graph embeddings and ontology-derived category features were progressively added to the baseline tabular models. Pairwise AUC comparisons showed that the neuro-symbolic configuration significantly outperformed the tabular baseline for all four model families: XGBoost ($p<0.001$), SVM ($p<0.001$), MLP ($p=0.046$), and Autoencoder ($p=0.012$). The strongest overall performance was achieved by XGBoost under the neuro-symbolic configuration, with an AUC of 0.77 (95\% CI: 0.76--0.79). Overall, these results show that incorporating ontology-based knowledge representations into our prediction process improves in-hospital mortality risk prediction.

\section{Explainability using SHAP and RAG Layer}
\begin{figure}[t]
  \centering
  \includegraphics[width=0.9\columnwidth]{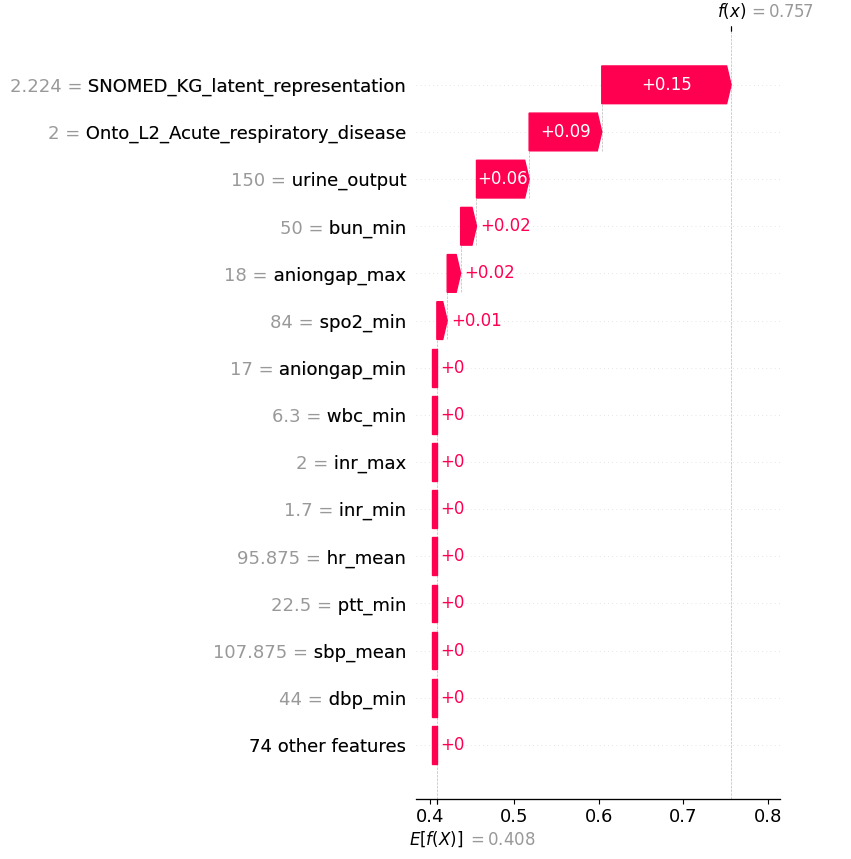}
  \caption{SHAP visualization - Neuro-symbolic MLP configuration. }
  \label{fig:shp_nls_output}
\end{figure}

For each experimental scenario, we performed SHAP analysis to visualize and quantify the influence of each variable on the mortality prediction outcome, thereby improving our understanding of the features contributing to mortality risk. The tree-based model, XGBoost, used TreeSHAP, while the MLP classifier, RBF SVM, and Autoencoder used KernelSHAP for this purpose \cite{lundbergUnifiedApproachInterpreting2017,lundberg2018consistent, shap_treeexplainer_doc}. SHAP was computed under the three feature configurations (tabular baseline, tabular + SNOMED graph embeddings, and full neuro-symbolic features). Figure~\ref{fig:shp_nls_output} shows a representative example of SHAP results from the MLP neuro-symbolic setting, in which the SNOMED embedding dimensions are collapsed into a single latent KG representation and visualized with Ontology category counts and tabular features. SHAP is a post hoc, model-agnostic XAI technique derived from game theory and based on Shapley values, which measure the average marginal contribution of a feature across all feature combinations. These values provide feature attribution scores for individual predictions, which, once aggregated, become feature importance scores that can be used to explain ML model predictions \cite{lundbergUnifiedApproachInterpreting2017, ortigossaEXplainableArtificialIntelligence2024}. Although this is a widely used explanation algorithm, owing to its ability to provide both local and global interpretability, studies suggest that it may be difficult for user groups to understand its output \cite{zytekExplingoExplainingAI2024, zytekLLMsXAIFuture2024}.  

\begin{table}[htbp]
\centering
\caption{Clinical thresholds for HF-related features.}
\label{tab:hf_feature_thresholds}
\footnotesize
\setlength{\tabcolsep}{3pt} 
\renewcommand{\arraystretch}{1.05} 

\begin{tabularx}{\columnwidth}{>{\raggedright\arraybackslash}p{1.60cm}
                              >{\raggedright\arraybackslash}X
                              >{\raggedright\arraybackslash}X
                              >{\raggedright\arraybackslash}p{2.00cm}}
\toprule
\textbf{Feature} &
\textbf{Author(s) substantiating feature use (HF)} &
\textbf{Author(s) substantiating threshold} &
\textbf{Threshold(s) supported in literature} \\
\midrule

Age &
Fonarow et al. \cite{fonarowRiskStratificationInHospital2005} &
Bauer et al. \cite{bauerCRB65PredictsDeath2006} &
$\ge$65 years \\

BUN &
Fonarow et al. \cite{fonarowRiskStratificationInHospital2005} &
Fonarow et al. \cite{fonarowRiskStratificationInHospital2005} &
$\ge$43 mg/dL \\

Urine output &
Testani et al. \cite{testaniRapidHighlyAccurate2016} &
Khwaja \cite{Notice2012} &
$<$0.5 mL/kg/h for $\ge$6 h \\

Anion gap (min/max) &
Samavarchitehrani et al. \cite{samavarchitehraniPrognosticValueAnion2024} &
Lou et al. \cite{louAssociationAniongap28day2024};
Huang et al. \cite{huangRelationshipAlbumincorrectedAnion2025} &
$\ge$16--18 mmol/L (high risk), $\ge$20 mmol/L (severe) \\

WBC (min) &
Bekwelem et al. \cite{bekwelemWhiteBloodCell2011}; &
Baddam \& Burns \cite{baddamSystemicInflammatoryResponse2025} &
$<$4 or $>$12 $\times 10^{9}$/L \\

Respiratory rate &
Andersen et al. \cite{andersen2016} &
Andersen et al. \cite{andersen2016}; Baddam \& Burns \cite{baddamSystemicInflammatoryResponse2025} &
$\ge$30 breaths/min \\

SpO$_2$ (min) &
Chioncel et al. \cite{chioncelPulmonaryOedemaTherapeutic2015} &
Smith et al. \cite{NationalEarlyWarninga} &
$\le$91 \% (highest NEWS2 band) \\

Heart rate (mean) &
B\"ohm et al. \cite{bohmHeartRateRisk2010a} &
Andersen et al. \cite{andersen2016} &
$\ge$130 bpm \\

SBP (mean) &
Fonarow et al. \cite{fonarowRiskStratificationInHospital2005}; &
Andersen et al. \cite{andersen2016} &
$\le$90 mmHg (abnormal), $\le$80 mmHg (severe) \\

DBP (min) &
Chioncel et al. \cite{chioncelPulmonaryOedemaTherapeutic2015} &
Zhao et al. \cite{zhaoImpactDiastolicBlood2025} &
$<$50--55 mmHg (critical hypotension) \\

INR (min/max) &
Ohgushi et al. \cite{ohgushiRiskMajorBleeding2016a} &
Ohgushi et al. \cite{ohgushiRiskMajorBleeding2016a} &
$\ge$3.0 \\

PTT (min) &
Levi \& Opal \cite{leviCoagulationAbnormalitiesCritically2006} &
Wang et al. \cite{wangDynamicChangesActivated2025} &
$>$37 s (prolonged); $\ge$50 s (severe prolongation) \\

Dobutamine &
Mebazaa et al. \cite{mebazaaLevosimendanVsDobutamine2007} &
Mebazaa et al. \cite{mebazaaLevosimendanVsDobutamine2007} &
Use vs no use \\

Dopamine &
De Backer et al. \cite{debackerComparisonDopamineNorepinephrine2010} &
De Backer et al. \cite{debackerComparisonDopamineNorepinephrine2010} &
Use vs no use \\

Norepinephrine &
De Backer et al. \cite{debackerComparisonDopamineNorepinephrine2010} &
De Backer et al. \cite{debackerComparisonDopamineNorepinephrine2010} &
Use vs no use \\

Phenylephrine &
Huang et al. \cite{huangRelationshipAlbumincorrectedAnion2025}; Lou et al.\cite{louAssociationAniongap28day2024} &
Huang et al. \cite{huangRelationshipAlbumincorrectedAnion2025}; Lou et al.\cite{louAssociationAniongap28day2024} &
Use vs no use \\

\bottomrule
\end{tabularx}
\end{table}
\subsection{Explanation Generation Pipeline}
The goal of this explanation generation pipeline is to enable the generation of explanations that are intelligible to a wider audience, including domain experts who may not routinely interpret ML model outputs. This layer translates existing machine learning explanations into clear, human-readable narratives grounded in established clinical knowledge. We first create a document corpus from the SNOMED ontology slice built for this AHF cohort, which converts concepts into text. We then create a Chroma collection and index the KG documents within it using Ollama embeddings (nomic-embed-text). This step enables us to ground our LLM in clinical knowledge and reduce hallucinations, since the system can now query the Chroma collection and retrieve stay-specific SNOMED evidence via a stay-level concept allow-list. 

A curated clinical knowledge base was subsequently added to provide directionality, typical thresholds, and brief interpretive statements for each of the 18 predictors, which were then sent to our RAG layer. Table~\ref{tab:hf_feature_thresholds} shows feature and threshold substantiation from existing literature used to curate this clinical knowledge base. We stored clinical notes in a persistent Chroma collection and retrieved them on a per-stay basis using metadata filtering on the ICU stay identifier to enhance patient-specific contextual grounding further. The SHAP outputs were transformed by ranking features by attribution magnitude and assigning directional effects based on their sign. Feature embeddings were collapsed into a single latent driver, since abstract embedding drivers cannot be directly mapped to existing features and do not contribute towards explainability. It enables predictive accuracy but not narrative clarity. Then we separated the features into three semantic groups (tabular, ontology, and embedding) to achieve structural separation and prevent knowledge leakage between the sections of our narrative. Driver tokens were used to ensure all high-impact contributors were covered in the generated narrative explanation. 

\begin{figure}[htbp]
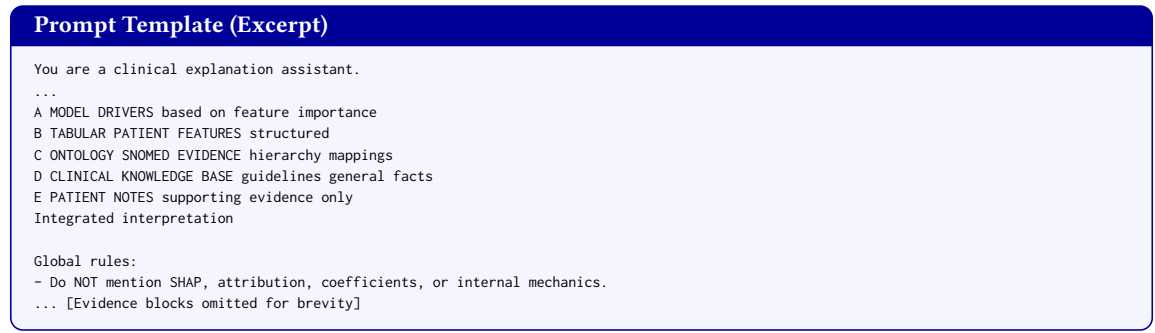

\centering
\begin{tcblisting}{
  colback=blue!4,
  colframe=blue!60!black,
  boxrule=0.6pt,
  arc=1.5mm,
  title={\textbf{Prompt Template (Excerpt)}},
  left=2mm,right=2mm,top=1mm,bottom=1mm,
  listing only,
  listing options={style=promptstyle}
}
You are a clinical explanation assistant.
... 
A MODEL DRIVERS based on feature importance 
B TABULAR PATIENT FEATURES structured 
C ONTOLOGY SNOMED EVIDENCE hierarchy mappings 
D CLINICAL KNOWLEDGE BASE guidelines general facts 
E PATIENT NOTES supporting evidence only 
Integrated interpretation 

Global rules: 
- Do NOT mention SHAP, attribution, coefficients, or internal mechanics.
...
Integrated interpretation:
...
- Explicitly state how the synthesized picture relates to the patient's overall mortality risk level.
... [Evidence blocks omitted for brevity]
\end{tcblisting}
\caption{Excerpt of the constrained prompt template.}
\label{fig:prompt_template}
\end{figure}

\begin{figure}[htbp]
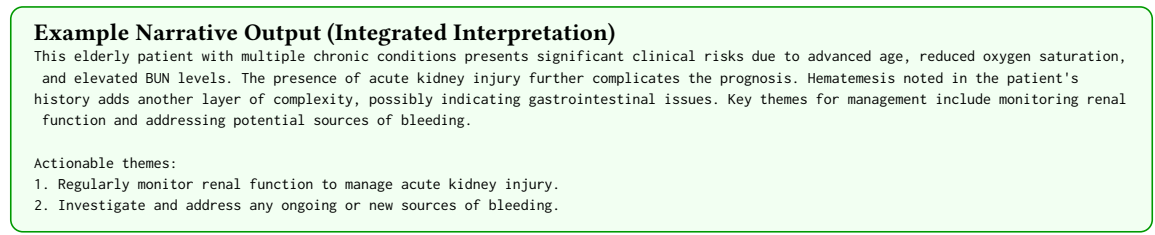

\centering
\begin{tcblisting}{
  colback=green!5,
  colframe=green!60!black,
  boxrule=0.6pt,
  arc=1.5mm,
  title={\textbf{Example Narrative Output (Integrated Interpretation)}},
  coltitle=black,
  colbacktitle=green!10,
  attach title to upper,
  fonttitle=\bfseries,
  left=2mm,right=2mm,top=1mm,bottom=1mm,
  listing only,
  listing options={style=promptstyle}
}
The patient exhibits multiple risk factors that collectively elevate their mortality risk. Elevated BUN levels and reduced urine output suggest cardiorenal syndrome and renal hypoperfusion, respectively. Additionally, low SpO2 indicates significant hypoxemia, likely due to pulmonary edema. The presence of acute respiratory failure further complicates the clinical picture. These findings highlight a critical need for vigilant monitoring and management strategies to address these severe conditions.

Actionable themes:
1. Intensify monitoring of renal function and fluid balance.
2. Enhance respiratory support measures to improve oxygenation levels. 
\end{tcblisting}
\caption{Representative integrated narrative explanation.}
\label{fig:example_output}
\end{figure}

A narrative explanation was generated using the RAG-grounded LLM by feeding SHAP-derived drivers, tabular features (along with threshold-derived status), a curated clinical knowledge base, ontology-related information, and ICU-specific stay notes as supporting evidence. The final prompt enforced a strict structure in which each context-based piece of evidence had its own section and an integrated synthesis. The sections that discuss SHAP-derived drivers and tabular features, along with their values, status, and effects, were generated deterministically using Python code and then fed into the prompt to reduce hallucination. Parts of the prompt, including the global constraints and the section-level rules, along with the generated explanation, are provided in Figures~\ref{fig:prompt_template} and~\ref{fig:example_output}. The final narrative explanation provides a more holistic picture of the patient's status.

\subsection{Explanation Evaluation Metrics}
The metrics we used to evaluate explanation quality between SHAP-generated and LLM-generated narrative texts were completeness, consistency, fidelity, robustness, and several human-aligned factors, such as perceived coherence, faithfulness, plausibility, usefulness, and sensemaking, assessed using the LLM-as-a-judge model. SHAP explanations were numeric vectors over the full neuro-symbolic framework without feature collapse. We used axiomatic, perturbation-based, and stability metrics grounded in attribution theory to evaluate SHAP explanations. The narrative text was evaluated for coverage of important features, consistency across multiple generations of the same explanation, and alignment with human criteria. This separation ensured that we were using metrics appropriate to each explanation modality while still comparing along the same lines. We also converted SHAP explanations into a text narrative with KG embedding dimensions collapsed into a single latent feature, to compare them with the narrative explanation using human-aligned (HA) metrics. Narrative explanations were generated using the model qwen2.5:14b-instruct, while evaluation was performed using a different model (gpt-oss:20b) to avoid self-evaluation bias. Table~\ref{tab:explainability_metrics_llm} summarizes the metrics used for SHAP and LLM-generated explanations, along with the evaluation methodology.

\begin{table*}[!htbp]
\footnotesize
\centering
\caption{Explainability metrics for SHAP and LLM-based measures with conceptual rationale and measurement approach.}
\label{tab:explainability_metrics_llm}
\setlength{\tabcolsep}{2pt}
\renewcommand{\arraystretch}{0.95}
\begin{tabularx}{\textwidth}{@{}
p{2.5cm}
>{\hsize=0.55\hsize}X
>{\hsize=1.45\hsize}X
@{}}
\toprule
\textbf{Metric} & \textbf{What it Measures} & \textbf{How it is Measured} \\
\midrule
\textbf{Completeness} 
& Whether the feature attributions fully account for the model’s prediction for a given instance.
& Quantified by the normalized residual between the sum of SHAP attributions and the model’s output difference from the baseline \cite{ higham2002accuracy,lundbergUnifiedApproachInterpreting2017,  shap_treeexplainer_doc}. 
We report a scaled diagnostic score:
\(\displaystyle
\text{CompletenessScore}(x)
=
1-\frac{\left|\sum_{j=1}^{d}\phi_j-(f(x)-\phi_0)\right|}
{|f(x)-\phi_0|+\varepsilon},
\qquad \varepsilon=10^{-12}.
 \)\\
\textbf{Additivity Gap} 
& Numerical deviation from SHAP’s completeness axiom.
& Computed as the absolute difference between the model output and the sum of SHAP attributions plus the baseline which serves as a diagnostic measure of deviation from the SHAP additivity condition \cite{lundbergUnifiedApproachInterpreting2017,shap_treeexplainer_doc}.

\(\displaystyle
\text{AdditivityGap}(x)
=
\left|f(x)-\left(\phi_0+\sum_{j=1}^{d}\phi_j\right)\right|,
\)\\

\textbf{Infidelity} 
& Whether feature attributions accurately predict changes in the model output under random masking perturbations.
& Measured by evaluating how well SHAP attributions predict output changes under random feature perturbations using Monte Carlo masking \cite{yehInfidelitySensitivityExplanations2019}.

\(\displaystyle
\mathrm{INFD}(\Phi,f,x)
=
\mathbb{E}_{I}
\left[
\left(
I^\top \phi
-
\bigl(f(x)-f(x_I)\bigr)
\right)^2
\right],
\)
\\

\textbf{Max-Sensitivity} 
& The maximum change of explanations under small input perturbations.
& Estimated by repeatedly perturbing the input within a bounded neighborhood and measuring the maximum change in SHAP attributions \cite{yehInfidelitySensitivityExplanations2019}.
\par\vspace{2pt}
\(\displaystyle
\mathrm{SENS}_{\max}(\Phi,f,x,r)
=
\max_{\|\delta\|\le r}
\left\|
\Phi(f,x+\delta)-\Phi(f,x)
\right\|_2,
\)
\\

\textbf{Robustness (Top-$k$ Jaccard)} 
& Stability of feature attribution under noisy inputs.
& Computed as the average Jaccard similarity between the top-$k$ SHAP features by comparing each perturbed explanation to the original reference explanation to get one-vs-reference robustness measure \cite{alvarez2018robustness, nogueiraStabilityFeatureSelection}.
\(\displaystyle
\mathrm{Robust}_k=\frac{1}{T}\sum_{t=1}^{T}
\frac{|S_k(x)\cap S_k(x^{(t)})|}{|S_k(x)\cup S_k(x^{(t)})|},
\quad
S_k(x)=\operatorname{TopK}(|\Phi(x)|).
\)
\\

\textbf{Run-to-Run Stability (Cosine)} 
& Reproducibility of SHAP attribution vectors across repeated SHAP runs on the same instance.
& Calculated as the average pairwise cosine similarity of normalized SHAP vectors across $R=30$ repeated runs on the same input, averaged across all pairs
 \cite{kelodjou_shaping_2024,levy_guide_2025}. 
 \par\vspace{2pt}
 \(\displaystyle
\mathrm{Stab}_{\cos}
=
\frac{1}{\binom{R}{2}}
\sum_{1 \le i < j \le R}
\frac{\phi^{(i)\top}\phi^{(j)}}{\|\phi^{(i)}\|_2\|\phi^{(j)}\|_2},
\qquad
\mathrm{Stab}_{\cos}^{[0,1]}=\frac{1+\mathrm{Stab}_{\cos}}{2}.
\)
 \\
\midrule
\textbf{SHAP-Mass Coverage} 
& Degree to which the narrative explanation covers model-identified important features.
& Computed as the proportion of top-K SHAP attribution mass corresponding to a set of features mentioned in the generated explanation denoted by $M$ \cite {deyoungERASERBenchmarkEvaluate2020,haseWhenCanModels2021,jacoviFaithfullyInterpretableNLP2020,zytekExplingoExplainingAI2024}.
\par\vspace{2pt}
\(\displaystyle
\text{MassCoverage}
=
\frac{
\sum_{j\in \text{TopK}(\mathcal{K}_{\text{filtered}})\cap M}|\phi_j|
}{
\sum_{j\in \text{TopK}(\mathcal{K}_{\text{filtered}})}|\phi_j|
}.
\)
Here, $K_{\text{filtered}}$ denotes the set of non-embedding features satisfying
$|\phi_j| \geq 0.01$, consistent with the threshold applied for narrative driver selection and SHAP text rendering.
\\

\textbf{Narrative Completeness (Explingo-based)} 
& Coverage of required driver elements (feature name, value, and direction) in the narrative.
& Evaluated as the fraction of required explanation items (feature name, value, and direction) correctly mentioned in the narrative
\cite{zytekExplingoExplainingAI2024}. 
For each required explanation item $i\in\{1,\dots,N\}$, we evaluate whether the narrative provides the feature name, associated value, and correct risk direction.
Individual coverage rates are:

\(\displaystyle
C_{\text{name}}=\frac{1}{N}\sum_{i=1}^{N}\mathbf{1}[\text{Name}_i],\quad
C_{\text{value}}=\frac{1}{N}\sum_{i=1}^{N}\mathbf{1}[\text{Value}_i],\quad
C_{\text{dir}}=\frac{1}{N}\sum_{i=1}^{N}\mathbf{1}[\text{Direction}_i].
\)

An item is fully complete only if all three criteria hold jointly; thus the final
Explingo-style completeness score (corresponding to $\texttt{all\_ok}/N$) is:

\(\displaystyle
C_{\text{Explingo}}
=
\frac{1}{N}\sum_{i=1}^{N}
\mathbf{1}\bigl[\text{Name}_i \wedge \text{Value}_i \wedge \text{Direction}_i\bigr].
\)
\\

\textbf{Run-to-Run Stability (Cosine)} 
& Reproducibility of feature mentions in LLM-generated narratives across multiple regenerations for the same instance.
& Calculated as the average pairwise cosine similarity of binary feature-mention vectors extracted from $R=30$ regenerated explanations, averaged across all pairs \cite{kelodjou_shaping_2024,levy_guide_2025}.
\par\vspace{2pt}
\(\displaystyle
\text{Stability}_{\cos}
=
\frac{1}{\binom{R}{2}}
\sum_{1 \le i < j \le R}
\frac{\mathbf{m}_i \cdot \mathbf{m}_j}
{\|\mathbf{m}_i\|_2 \, \|\mathbf{m}_j\|_2},
\qquad
\text{Stability}_{\cos}^{[0,1]}
=
\frac{1 + \text{Stability}_{\cos}}{2}.
\)
\par\vspace{2pt}
where $\mathbf{m}_i \in \{0,1\}^d$ is the binary feature-mention vector extracted from the $i$-th narrative regeneration, $d$ is the number of features.
\\

\textbf{Clinical Plausibility} 
& Consistency of explanations with established clinical knowledge and guidelines.
& Assessed by comparing narrative claims against curated clinical thresholds, expected risk directionality, and guideline-based rules
\cite{cabitzaUnintendedConsequencesMachine2017,doshi-velezRigorousScienceInterpretable2017a,holzingerCausabilityExplainabilityArtificial2019,tonekaboniWhatCliniciansWant2019}. 
\par\vspace{2pt}
\(\displaystyle
C_{\text{plaus}}
=
\frac{1}{m}\sum_{j=1}^{m}s_j,
\qquad
s_j\in\{1.0,\,0.6,\,0.5,\,0.0\},
\)
\par\vspace{2pt}
where $s_j$ reflects agreement (1.0), partial credit (0.6), neutral mention (0.5), or contradiction (0.0). 
\\

\textbf{Human-Aligned Faithfulness} 
& Perceived correctness of explanations relative to the model’s reasoning.
& Assessed using an LLM-as-a-judge that rates whether the explanation reflects the model inputs/contributors provided, without inventing unsupported cause
\cite{badshahReferenceGuidedVerdictLLMsasJudges2025,guSurveyLLMasaJudge2025,hoffmanEvaluatingMachinegeneratedExplanations2023,kadirEvaluationMetricsXAI2023a}.\\

\textbf{Human-Aligned Plausibility} 
& Clinical and logical reasonableness of explanations.
& Assessed using an LLM-as-a-judge that rates whether a clinician would find the explanation medically sensible and coherent \cite{badshahReferenceGuidedVerdictLLMsasJudges2025,guSurveyLLMasaJudge2025,hoffmanEvaluatingMachinegeneratedExplanations2023,kadirEvaluationMetricsXAI2023a}. \\

\textbf{Human-Aligned Usefulness} 
& Practical utility of explanations for decision support.
& Assessed using an LLM-as-a-judge that rates whether explanations help users understand, anticipate, or act on model predictions
\cite{badshahReferenceGuidedVerdictLLMsasJudges2025,guSurveyLLMasaJudge2025,kadirEvaluationMetricsXAI2023a,mohseniQuantitativeEvaluationMachine2021, nautaAnecdotalEvidenceQuantitative2022a}.\\

\textbf{Human-Aligned Sensemaking} 
& Perceived sensemaking of explanations.
& Assessed using an LLM-as-a-judge that rates whether explanations support clinical sensemaking by identifying salient features, organizing them, and constructing an interpretive narrative without overstating or hallucinating
 \cite{hoffmanEvaluatingMachinegeneratedExplanations2023,article}. \\

\textbf{Human-Aligned Overall} & 
Overall quality of the explanation as judged by the LLM evaluator, considering faithfulness, plausibility, usefulness, and sensemaking. & 
Computed as the unweighted mean of the four human-alignment sub-dimension scores (Faithfulness, Plausibility, Usefulness, Sensemaking). \\

\textbf{Narrative Coherence} 
& Internal consistency, clarity, and structured flow.
& Assessed using an LLM-as-a-judge that rates the explanation’s internal consistency, logical ordering, and absence of contradictions on a continuous scale from 0 to 1.
\cite{badshahReferenceGuidedVerdictLLMsasJudges2025,guSurveyLLMasaJudge2025,kadirEvaluationMetricsXAI2023a, nautaAnecdotalEvidenceQuantitative2022a}. If the judge output could not be parsed as a numeric score, a deterministic text-length/keyword heuristic was substituted. \\
\bottomrule
\end{tabularx}
\end{table*}
\section{Explanation Quality Comparison Results}
The metric calculations were run 30 times to obtain aggregate metrics, rather than relying on a single run that could be compromised by cherry-picking "good" or "bad" explanations. Across the repeated regenerations, the SHAP vector, stay data, and retrieved clinical evidence were held fixed for the LLM narratives, whereas the SHAP text used resampled attribution vectors. SHAP-Mass Coverage and Narrative Completeness for each LLM narrative were evaluated against the corresponding resampled vector. The resampled attribution vectors showed high similarity to the fixed SHAP vector (mean cosine similarity = 0.96 across 30 runs). Explainability metrics are reported in Table~\ref{tab:shap_rag_fullwidth} as means with standard deviations and 95\% confidence intervals. SHAP results show near-exact completeness (1.000 ± 0.000) and a negligible additivity gap (0.000 ± 0.000), consistent with SHAP's local-accuracy axiom. However, SHAP performed worse on the human-aligned criteria (Overall = 0.558; Sensemaking = 0.270), indicating limited clinical interpretability. In contrast, LLM-generated narrative explanations achieved higher human-aligned performance (Overall = 0.807; Sensemaking = 0.793), indicating excellent clinical interpretability, while maintaining high coverage of SHAP model drivers (SHAP-Mass coverage = 0.942). SHAP stability was computed over the full, non-feature-collapsed attribution vectors, while narrative stability reflects the consistency of clinical factor mentions rather than the generation of identical text. SHAP-side human-aligned faithfulness scores may partially reflect run-to-run variability. Run-to-Run Stability for narratives reached a ceiling, since the mentioned feature set was essentially unchanged. 
\begin{table}[!htbp]
\centering
\caption{Explanation evaluation metrics across 30 repeated runs.}
\label{tab:shap_rag_fullwidth}
\footnotesize
\setlength{\tabcolsep}{3pt}
\renewcommand{\arraystretch}{1.10}
\begin{tabular}{p{3.4cm}p{2.0cm}p{2.0cm}}
\hline
\textbf{Metric} & \textbf{SHAP} & \textbf{RAG-Grounded Narrative} \\
\hline
Completeness & 1.000 $\pm$ 0.000 & -- \\
 & (1.000--1.000) & \\
Additivity Gap & 0.000 $\pm$ 0.000 & -- \\
 & (0.000--0.000) & \\
Infidelity & 0.022 $\pm$ 0.001 & -- \\
 & (0.022--0.022) & \\
Max-Sensitivity & 0.042 $\pm$ 0.019 & -- \\
 & (0.035--0.049) & \\
Robustness (Top-k Jaccard) & 0.748 $\pm$ 0.104 & -- \\
 & (0.711--0.785) & \\
\hline
SHAP-Mass Coverage & -- & 0.942 $\pm$ 0.047 \\
 & & (0.925--0.959) \\
Narrative Completeness & -- & 0.718 $\pm$ 0.148 \\
 & & (0.665--0.771) \\
Clinical Plausibility & -- & 0.683 $\pm$ 0.002 \\
 & & (0.682--0.683) \\
\hline
HA Faithfulness & 0.753 $\pm$ 0.171 & 0.761 $\pm$ 0.138 \\
 & (0.692--0.814) & (0.712--0.811) \\
HA Plausibility & 0.707 $\pm$ 0.139 & 0.852 $\pm$ 0.071 \\
 & (0.657--0.756) & (0.826--0.877) \\
HA Usefulness & 0.502 $\pm$ 0.114 & 0.823 $\pm$ 0.078 \\
 & (0.461--0.542) & (0.795--0.851) \\
HA Sensemaking & 0.270 $\pm$ 0.063 & 0.793 $\pm$ 0.107 \\
 & (0.247--0.293) & (0.754--0.831) \\
HA Overall & 0.558 $\pm$ 0.098 & 0.807 $\pm$ 0.091 \\
 & (0.523-0.593) & (0.775-0.840) \\
\hline
Narrative Coherence & -- & 0.801 $\pm$ 0.083 \\
 & & (0.771--0.831) \\
Run-to-Run Stability (Cosine) & 0.971 & 1.000 \\
\hline
\end{tabular}

\footnotesize
\vspace{0.3em}
 Values are mean $\pm$ SD with 95\% confidence intervals in parentheses. \textbf{HA} = Human-Aligned. Dashes indicate metrics not applicable. 
\end{table}

\section{Challenges and Limitations}
We encountered several challenges in producing the narrative text. The prompt design required several iterations to minimize information leakage and ensure that the LLM did not invent unsupported clinical knowledge. The final integration paragraph was also challenging to produce and constrain, particularly to maintain narrative prose while remaining useful and actionable for clinicians. Despite explicit rules and five distinct knowledge sources, the LLM still hallucinated and introduced extraneous external information into the narrative that could not be derived from the pipeline input. As a result, portions of the prompt context required sanitization and abstraction to improve interpretability, rather than relying on the raw feature text. Because the output failed to adhere to the rules outlined in the prompts in multiple instances, we employed strict, rule-based prompting with global and section-specific validity requirements. We observed that the narrative output was highly dependent on the LLM used, and prompts provided require careful grounding and evaluation to prevent hallucination or overgeneralization. A smaller locally hosted model, such as gemma3:4b (4 GB), exhibited frequent hallucinations and reduced coherence. In contrast, a larger locally hosted model produced more reliable results (qwen2.5:14b-instruct (9 GB) for explanation generation and gpt-oss:20b (12 GB) for LLM-as-a-judge evaluation). In some runs with gemma3:4b, the output still indicated that the patient was receiving "Dobutamine" and "Dopamine," even though they were not.  The rules in the prompt clearly stated that these were binary features (value = 1, drug administered; value = 0, drug not administered). This may be due to the heterogeneous nature of data (notes, clinical knowledge, SHAP results) passed to the LLM layer, which requires deep contextual reasoning, and hence a model with a higher number of parameters generates better results. Furthermore, human-aligned evaluation dimensions rely on the LLM-as-a-judge framework, which might introduce model-dependent bias.

\section{Conclusion}
Our results indicate that predictive performance gains from ontology integration are not model-specific, as they are generally observed across all four models to which we added SNOMED embeddings and ontological features. While AUC consistently improved with neuro-symbolic integration, gains in sensitivity and F1 score were more variable across model families. We observe that SHAP-generated output excels in run-to-run stability (cosine similarity of attribution vectors), and exhibits relatively high robustness of top-k feature attribution under perturbation. SHAP's cosine similarity scores may be driven by a few features with large magnitudes across runs. Robustness score, by contrast, indicates some variation in SHAP's top-k feature ranking under small input changes. LLM-generated narrative output exhibits high narrative coherence and feature-mention stability, synthesizing multiple evidence sources, including unstructured clinical notes, SHAP-derived feature-attributions, ontology context, and curated clinical knowledge to provide a holistic picture. These narratives outperform SHAP on human-aligned metrics such as faithfulness, plausibility, usefulness, and sensemaking. The results highlight a trade-off: both explanations are valuable in different ways and could be combined to produce outputs that address a wider audience.

The proposed NEURON framework yields improved predictive performance. It produces an explanatory narrative, steeped in clinical knowledge and patient-specific context, expressed in natural language, to ensure that different providers can gain a holistic understanding of mortality risk. Although the narrative explanation exhibits high coherence and strong human-aligned metrics when evaluated with an LLM-as-a-judge, it should be used as complementary to existing XAI techniques, such as SHAP, because it can still produce hallucinations and is dependent on the models used to generate and evaluate the explanations. 

\section{Future Work}
For future work, we aim to apply NEURON to other diagnostic domains, such as readmissions. We will work on scaling NEURON to support other clinical applications, including expanding the clinical knowledge base. Our future work will also focus on understanding the propagation of instability across multiple layers of our NEURON framework, including the SHAP attribution and the narrative generation layers. Our next step is to conduct a human assessment of this framework, considering the impact that AI systems have on patient safety. We plan to implement this via a mixed-methods evaluation with clinicians and healthcare professionals using surveys. This will enable the development of standardized, human-centered metrics to evaluate explanation quality and further assess perceptions of trust and risk based on the generated explanations. A systematic comparison across various LLM models will be valuable for identifying models that generate narratives closest to human understanding while remaining grounded in clinical knowledge. Finally, an important next step is to transition from static narrative explanations to an interactive reasoning system that supports clinical dialogue in real-world workflows.

\section{Acknowledgement}
This work was supported in part by the National Science Foundation (NSF) under Award 2505686.

\bibliographystyle{ACM-Reference-Format}
\bibliography{bibfile}

\end{document}